\pdfoutput=1

\documentclass[11pt]{article}

\usepackage[preprint]{acl}

\usepackage{times}
\usepackage{latexsym}
\usepackage[T1]{fontenc}
\usepackage[utf8]{inputenc}
\usepackage{inconsolata}
\usepackage{CJKutf8}

\usepackage{microtype}
\usepackage{lscape}

\usepackage{amsmath}
\usepackage{amssymb}
\usepackage{amsfonts}
\usepackage{amsthm}
\usepackage{nicefrac}

\usepackage{booktabs}
\usepackage{multirow}
\usepackage{colortbl}
\usepackage{threeparttable}

\usepackage{graphicx}
\usepackage{float}
\usepackage{placeins}
\usepackage{subcaption}

\usepackage[dvipsnames]{xcolor}
\definecolor{mygray}{gray}{.88}
\definecolor{pyblue}{rgb}{0.0, 0.0, 0.5}
\definecolor{pygreen}{rgb}{0.0, 0.5, 0.0}
\definecolor{pyorange}{rgb}{1.0, 0.4, 0.0}
\definecolor{myblue}{HTML}{5f8fb7}
\definecolor{myred}{HTML}{cc7c7c}
\definecolor{mybrown}{HTML}{D8C6A6}
\definecolor{mossgreen}{HTML}{8B9A7B}

\usepackage{enumitem}

\usepackage{algorithm}
\usepackage{algpseudocode}
\usepackage{listings}
\usepackage{alltt}

\lstdefinestyle{pythonstyle}{
    language=Python,
    basicstyle=\footnotesize\ttfamily,
    breaklines=true,
    morekeywords={self},
    keywordstyle=\color{pyblue},
    commentstyle=\color{pygreen},
    stringstyle=\color{pyorange},
    numberstyle=\tiny\color{gray},
    numbers=left,
    numbersep=10pt,
    tabsize=4,
    showspaces=false,
    showstringspaces=false
}

\usepackage[most]{tcolorbox}
\tcbuselibrary{breakable}

\tcbuselibrary{skins,breakable}
\usepackage{seqsplit}

\usepackage{pifont}
\usepackage{fontawesome5}

\usepackage{tikz}

\usepackage{url}
\usepackage{cleveref}

\usepackage{xspace}
\usepackage{fancyvrb}
\usepackage{fvextra} 
\usepackage{titletoc}

\definecolor{lightgray}{gray}{0.9} 
\definecolor{lightergray}{gray}{0.95} 

\usepackage{dsfont} 

\usepackage{amsmath,amsfonts,bm}

\def\eqref#1{equation~\ref{#1}}

\def\1{\bm{1}}

\DeclareMathAlphabet{\mathsfit}{\encodingdefault}{\sfdefault}{m}{sl}
\SetMathAlphabet{\mathsfit}{bold}{\encodingdefault}{\sfdefault}{bx}{n}

\newlength\savewidth

\definecolor{plmcolor}{HTML}{D6FAE1}    
\definecolor{plmtag}{HTML}{2EAA85}
\definecolor{llmpcolor}{HTML}{D6F6FD}   
\definecolor{llmptag}{HTML}{3498DB}
\definecolor{llmtcolor}{HTML}{FDECDF}   
\definecolor{llmttag}{HTML}{E86B5D}
\definecolor{gold}{HTML}{FFD700}
\definecolor{silver}{HTML}{C0C0C0}
\definecolor{bronze}{HTML}{CD7F32}

\newcommand{\method}{{\fontfamily{lmtt}\selectfont \textbf{PHI}}\xspace}
\newcommand{\llmname}[1]{{\fontfamily{lmtt}\selectfont #1}}

\title{Can Post-Training Turn LLMs into Good Medical Coders? \\ An Empirical Study of Generative ICD Coding}

\author{
  Ziqing Wang, \ Weihao Li, \ Shijie Chen, \ Yuan Luo, \ Kaize Ding \\
  Northwestern University \\}

\begin{document}
\maketitle
\begin{abstract}
Automated International Classification of Diseases (ICD) coding is a core medical-coding task for billing, epidemiology, and clinical decision support. Generative large language models (LLMs) are often reported as weak medical coders, but this finding mainly comes from inference-time settings such as prompting, retrieval, reranking, or tool use, leaving the role of task-specific post-training underexplored. We present a controlled empirical study of post-training for generative ICD coding, comparing discriminative baselines with LLM coders across prompting, supervised fine-tuning, and reinforcement learning under a common protocol and metric set. To our knowledge, this is the first study to evaluate RL-based post-training for generative LLM coders in ICD coding. We further introduce \method{}, a diagnostic curriculum that extends GRPO to refine missed-code cases. Our results show that prompting-only evaluation substantially underestimates the potential of LLMs for ICD coding. SFT provides the main capability jump, GRPO further improves code-set prediction beyond SFT, and \method{} provides targeted gains on macro-level performance. These findings suggest that the main bottleneck is not the generative formulation alone, but how the model is adapted and optimized for full-taxonomy recall. We release our code, data splits, and checkpoints at \url{https://github.com/AlexandreWANG915/LLM4ICD}.
\end{abstract}    
\section{Introduction}
\label{sec:intro}

Automated International Classification of Diseases (ICD) coding, a central form of medical coding, aims to map each clinical note to a set of standardized diagnosis and procedure codes~\citep{mullenbach2018explainable,teng2022review,ji2024unified}. These codes are widely used for medical billing, epidemiology, and clinical decision support~\citep{teng2022review,ji2024unified}. The task is difficult because a single discharge summary can span thousands of tokens and must be mapped to a subset of an enormous taxonomy, with over 70,000 codes in ICD-10-CM~\citep{mullenbach2018explainable,vu2020label}. As a result, ICD coding is an extreme multi-label problem with a highly imbalanced label distribution, making accurate code prediction a major challenge~\citep{edin2023automated}.

\begin{figure}[t]
\centering
\includegraphics[width=\linewidth]{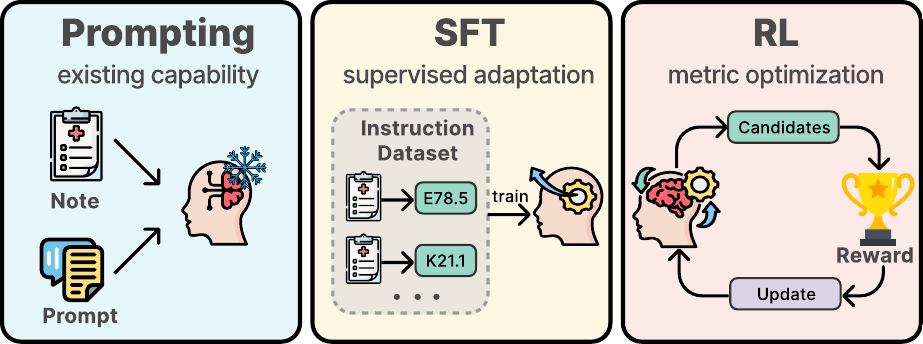}
\caption{\textbf{From prompting to post-training for ICD coding.} Prompting relies on an LLM's existing inference-time capability. SFT adapts the model from labeled note-code demonstrations, teaching a parseable output schema and an empirical code prior. RL further optimizes generated candidate code sets with a sample-level F1 reward computed from the parsed codes.}
\label{fig:posttraining-teaser}
\vspace{-0.3cm}
\end{figure}

Prior work has mainly addressed this challenge with discriminative ICD coders. Built on pretrained language model (PLM) encoders and label-wise prediction heads, these systems score codes in a fixed ICD label space and remain strong baselines for long-document coding~\citep{mullenbach2018explainable,vu2020label,huang2022plm,edin2023automated}. In contrast, generative LLMs have often been reported as weak medical coders~\citep{soroush2024large}. Most existing generative studies evaluate LLMs through inference-time use, including zero-shot or few-shot prompting, chain-of-thought reasoning, retrieval, reranking, or tool use, rather than adapting the model itself for ICD coding~\citep{soroush2024large,boyle2023automated,kwan2024large,baksi2025medcoder}. These settings often lead to hallucinated or invalid codes and poor exact-code performance, but they leave open whether the same generative models can become accurate coders after task-specific post-training. This distinction matters because the generative formulation remains attractive: LLMs provide a natural-language interface, can follow task instructions, and can emit ICD code sets directly as text.

Task-specific post-training offers a natural path forward, since it adapts the model itself rather than relying only on inference-time prompting or tools. In recent LLM development, supervised fine-tuning (SFT) has become a standard first stage for this adaptation: the model is trained on instruction-response pairs so that it learns the task format, domain style, and output conventions~\citep{ouyang2022training,wei2021finetuned}. For ICD coding, SFT can teach the model to produce valid code lists in the required format while learning the empirical distribution of medical codes. However, SFT is still trained by maximum likelihood. It optimizes next-token prediction rather than the non-differentiable set-level metrics, such as precision, recall, and F1, that determine coding performance~\citep{ranzato2015sequence}. Reinforcement learning (RL) provides a complementary post-training stage: a complete model output is scored by a reward function, and the policy is updated to increase outputs with higher reward. This reward-based view has proven useful in modern LLM post-training, from human-preference alignment to mathematical reasoning and other verifiable tasks~\citep{schulman2017proximal,ouyang2022training}. ICD coding is a natural fit for this framework because each generated code set can be scored directly by an F1-based reward. Yet RL-based post-training for generative LLM coders in ICD coding remains unexplored.

We address this gap through a controlled empirical progression across two MIMIC datasets, two ICD code systems, Top-50 and Full label settings, and multiple LLM backbones. As illustrated in Figure~\ref{fig:posttraining-teaser}, we organize the study as a staged post-training ladder: \ding{182}\ \textbf{SFT} establishes the output schema and empirical code prior; \ding{183}\ \textbf{GRPO} uses a sample-level F1 reward to optimize generated code sets; and \ding{184}\ \textbf{\method{}} (Progressive Hint Injection) extends GRPO with a diagnostic missed-code curriculum, using codes missed by earlier checkpoints as stochastic training-time hints while keeping inference hint-free. To our knowledge, this is the first study to apply RL-based post-training to generative LLM coders for ICD coding. Across this progression, we find that prompting-only evaluation substantially underestimates the potential of LLMs for ICD coding. SFT provides the main capability jump, GRPO further improves code-set prediction beyond SFT, especially in Full label settings, and \method{} provides targeted gains on remaining missed-code cases and macro-level performance. Our contributions are:

\begin{itemize}[leftmargin=*,itemsep=-0.3pt]
    \item \textbf{Empirical reframing of generative ICD coding.} We show that prompting-only evaluation substantially underestimates the potential of generative LLMs for ICD coding. Under task-specific post-training, the conclusion changes from near-unusable prompting performance to competitive code-set prediction under controlled evaluation. We release code, data splits, and checkpoints to support reproducible comparison.

    \item \textbf{First RL-based post-training study for generative ICD coding.} To our knowledge, this is the first study to apply GRPO-style reinforcement learning to post-train generative LLMs for ICD code-set prediction. All generative methods use the same datasets, splits, parser, and metric set.

    \item \textbf{Diagnostic curriculum for missed-code recall.} We introduce \method{}, a training-time missed-code curriculum that extends GRPO with codes missed by earlier checkpoints while keeping inference hint-free, providing targeted refinement for remaining missed-code cases.
\end{itemize}
\section{Related Work}
\label{sec:related_work}

\paragraph{Discriminative ICD coding.} Automated ICD coding has traditionally been formulated as extreme multi-label classification over long clinical documents. CAML introduced code-specific attention to connect each ICD prediction with supporting spans in the note~\citep{mullenbach2018explainable}. Later discriminative models improved either the document encoder or the label representation: MultiResCNN uses multi-filter residual convolutions for long notes~\citep{li2020icd}, LAAT applies label attention with hierarchical learning for infrequent codes~\citep{vu2020label}, convolutional attention models target long-tailed clinical document classification~\citep{liu2021effective}, and label-correlation rerankers model dependencies among ICD codes~\citep{tsai2021modeling}. Encoder-based pretrained language models further strengthen this paradigm, with BERT-XML and PLM-ICD adapting contextual encoders to large ICD label spaces and long inputs~\citep{zhang2020bert,huang2022plm}. \citet{edin2023automated} show that these systems remain strong baselines on clean MIMIC-III and MIMIC-IV splits when preprocessing and thresholding are controlled. Our work uses these discriminative coders as strong reference points for evaluating whether task-specific post-training can make generative LLM coders competitive under the same protocol.

\paragraph{Generative ICD coding.}
Generative ICD coding treats code assignment as a text-generation problem: an LLM reads a clinical note and emits the applicable ICD codes as text. Existing work has mainly explored this formulation through inference-time use of the model, including zero-shot or few-shot prompting, chain-of-thought reasoning, retrieval, reranking, and tool use~\citep{boyle2023automated,soroush2024large,kwan2024large,baksi2025medcoder}. These studies show that generative coders can be flexible, but they often struggle with exact medical-code selection and can produce invalid or hallucinated codes~\citep{soroush2024large}. Other work has explored domain-specific fine-tuning or rationale-based supervision for improving generative medical coding~\citep{hou2025enhancing,li2026evaluation}. RL has also been applied to ICD coding, but prior work uses it for multi-agent path search over the ICD hierarchy with a discriminative policy network rather than to post-train a generative LLM~\citep{lu2025himarl}. Our work focuses on this missing post-training axis by evaluating how prompting, SFT, GRPO, and \method{} change generative ICD coding under a controlled protocol.

\paragraph{Post-training methods.}
Post-training has become a standard stage for adapting pretrained language models to downstream tasks. Supervised fine-tuning (SFT) trains models on instruction-response pairs and is commonly used to teach task format, domain style, and output conventions~\citep{ouyang2022training,wei2021finetuned}. However, maximum-likelihood training optimizes next-token prediction rather than non-differentiable task metrics, creating a mismatch between token-level learning and sequence- or set-level evaluation~\citep{ranzato2015sequence}. Reinforcement-learning-based post-training addresses this mismatch by scoring complete model outputs with explicit rewards and updating the policy toward higher-reward generations. PPO is widely used in RLHF because its clipped policy objective stabilizes updates while a KL penalty keeps the policy close to a reference model~\citep{schulman2017proximal,ouyang2022training}. More recent methods such as GRPO remove the learned value model and estimate advantages from groups of sampled responses, making them attractive for metric-driven post-training~\citep{shao2024deepseekmath}. Medical NLP has also studied domain and task adaptation for biomedical language models~\citep{alsentzer2019publicly,gu2021domain}. For generative ICD coding, however, post-training remains underexplored: existing work has mainly evaluated inference-time prompting, tool-augmented coding pipelines, or domain-specific fine-tuning, rather than a staged comparison of SFT and RL-based post-training under the same setup. Our work brings this post-training perspective to ICD coding by comparing SFT and GRPO with the same backbone, data, and evaluation protocol, then extending GRPO with \method{}.
\begin{figure*}[t!]
\centering
\includegraphics[width=\linewidth]{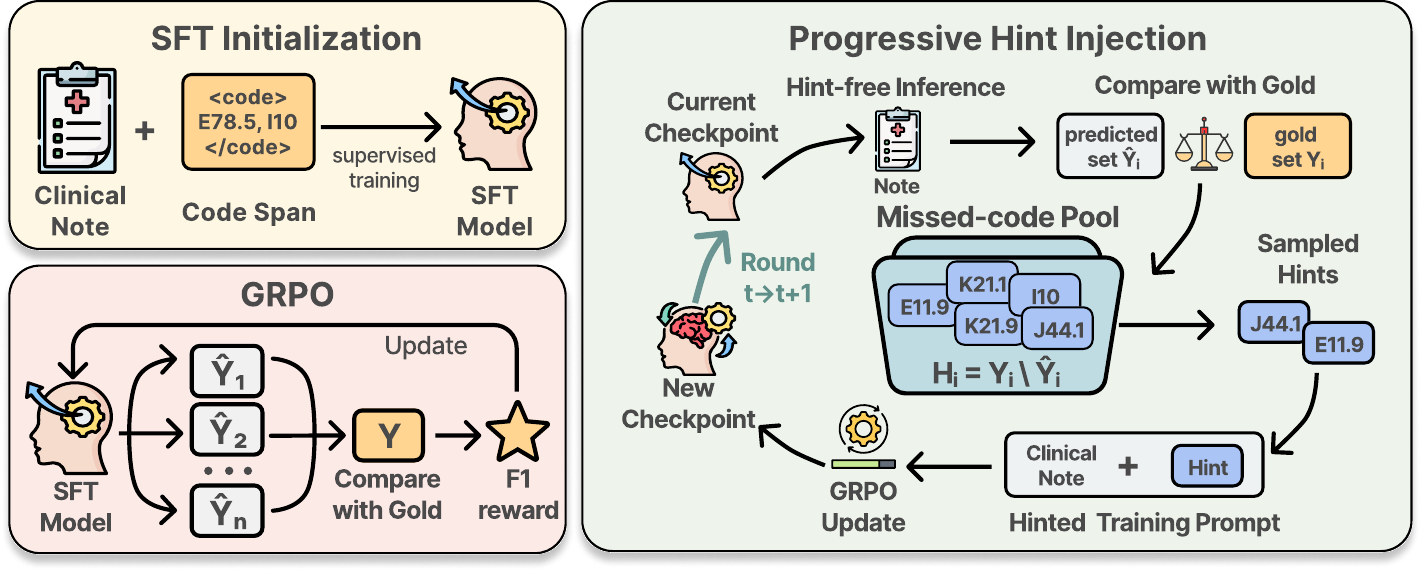}
\caption{\textbf{Overview of our post-training pipeline.} A generative LLM is first supervised fine-tuned to emit a parseable code span, then optimized with GRPO using a sample-level F1 reward computed from the parsed code set. Progressive Hint Injection (\method{}) iterates over rounds: it runs the current checkpoint without hints, collects missed codes \(Y_i \setminus \hat{Y}_i\) into a per-sample hint pool, and samples training-time hints from this pool for the next GRPO round. Hints are used only during training. At inference, the model receives only the clinical note.}
\vspace{-0.1cm}
\label{fig:pipeline}
\end{figure*}

\section{Preliminaries}
\label{sec:preliminary}

\paragraph{Task formulation.}
Given a clinical note $\mathbf{x}=(x_1,x_2,\ldots,x_n)$ and a predefined ICD code taxonomy $\mathcal{C}=\{c_1,c_2,\ldots,c_L\}$, automated ICD coding aims to predict the subset of applicable codes $Y\subseteq\mathcal{C}$. Equivalently, $Y$ can be represented by a binary vector $\mathbf{y}\in\{0,1\}^L$, where $y_j=1$ indicates that code $c_j$ applies to the note. We write the coding function abstractly as
\[
f:\mathbf{x}\mapsto \hat{Y}, \qquad \hat{Y}\subseteq\mathcal{C},
\]
where $\hat{Y}$ is the predicted code set. Since $\mathcal{C}$ can contain thousands of diagnosis and procedure codes, ICD coding is commonly treated as an extreme multi-label prediction problem.

Discriminative coders instantiate $f$ by assigning a score to each code in the fixed taxonomy and selecting positive labels with a validation-tuned threshold~\citep{mullenbach2018explainable,huang2022plm}. Generative coders instead instantiate $f$ through conditional text generation: given $\mathbf{x}$, an LLM autoregressively produces a structured response containing ICD codes, such as a \texttt{<code>...</code>} span, which is parsed into $\hat{Y}$. We use the same deterministic parser for all generative methods. Duplicate codes are merged, and malformed or invalid codes are excluded from $\hat{Y}$. All post-training methods in this paper operate on this generative formulation, differing only in how the LLM is adapted before producing the parsed code set.
\section{Method}
\label{sec:method}

We post-train a generative LLM for ICD coding through a staged pipeline (Figure~\ref{fig:pipeline}). First, supervised fine-tuning (SFT) teaches the model to emit a parseable code span and provides an empirical code prior (Section~\ref{sec:method:sft}). Second, GRPO optimizes the SFT policy with a sample-level F1 reward computed from the parsed code set (Section~\ref{sec:method:rl}). Third, Progressive Hint Injection (\method{}) extends GRPO with an iterative curriculum that focuses training on codes missed by earlier checkpoints (Section~\ref{sec:method:phi}). All stages output a single \texttt{<code>$c_1, c_2, \ldots, c_n$</code>} span, which we parse into the predicted set $\hat{Y}$. Hints are used only during training. At inference, the model receives only the clinical note, matching the hint-free setting used by all generative baselines.

\subsection{SFT Initialization}
\label{sec:method:sft}

Prompting relies on the base model's existing capability, but ICD coding requires a constrained output schema and an empirical code prior. We therefore use SFT as the first post-training stage, turning the instruction-tuned base model into a parseable generative ICD coder before RL.

Each SFT example consists of a user prompt $\mathbf{x}_i$, containing a fixed coding instruction, the discharge summary, and the required output format, paired with an assistant response $\mathbf{z}_i$ that serializes the gold code set $Y_i$ inside a single \texttt{<code>...</code>} span. Let $\mathcal{D}_{\mathrm{SFT}}=\{(\mathbf{x}_i,\mathbf{z}_i)\}_{i=1}^{N}$ denote these note-to-code demonstrations. We optimize the masked next-token objective
\[
\mathcal{L}_{\mathrm{SFT}}(\theta)
=
-\sum_{i=1}^{N}\sum_{t=1}^{|\mathbf{z}_i|}
\log \pi_{\theta}(z_{i,t}\mid \mathbf{x}_i,\mathbf{z}_{i,<t}).
\]
The loss is computed only over assistant response tokens, while user-prompt tokens are masked and do not contribute to the gradient. We implement this adaptation with LoRA and use the resulting policy $\pi_{\theta_{\mathrm{SFT}}}$ to initialize both GRPO and \method{}.

\subsection{GRPO with a Sample-Level F1 Reward}
\label{sec:method:rl}

SFT provides a valid generative coder, but its maximum-likelihood objective still trains the model at the token level. ICD coding, however, is evaluated after the generated text is parsed into a code set. We therefore use RL as a second post-training stage: each complete response is parsed into $\hat{Y}$, scored against the gold set $Y$, and used to update the generative policy. This lets the model optimize the quality of the predicted code set rather than only imitate the gold code string.

\paragraph{Reward.}
For each sampled response, we parse the predicted code set $\hat{Y}$ and compare it with the gold set $Y$. We use the sample-level F1 as the reward:
\[
    P = \frac{|\hat{Y} \cap Y|}{|\hat{Y}|}, \quad
    R = \frac{|\hat{Y} \cap Y|}{|Y|}, \quad
    r_{\mathrm{F1}} = \frac{2 P R}{P + R}.
\]
For malformed responses or predictions with no valid code span, we set $r_{\mathrm{F1}}=0$. Our main evaluation reports corpus-level micro-F1, but this metric is aggregated over many examples and is not suitable as a rollout reward. GRPO needs response-level rewards to compare sampled outputs within each group. We therefore use sample-level F1 as a per-response proxy that rewards the same precision--recall tradeoff over parsed code sets, while avoiding token-level likelihood alone.

\paragraph{Optimization.}
Starting from the SFT policy $\pi_{\theta_{\mathrm{SFT}}}$, we optimize the model with GRPO~\citep{shao2024deepseekmath}. GRPO is well suited to our setting because the reward is programmatically computed from the parsed code set, so no separate reward model is needed. It also avoids training a value critic by estimating relative advantages within a group of sampled responses, making it practical for RL post-training on large ICD datasets.

For each prompt $\mathbf{x}$, we sample a group of $G$ responses $\{o_1,\ldots,o_G\}$ from the old policy $\pi_{\theta_{\mathrm{old}}}$. Each response is parsed into a code set and scored with the sample-level F1 reward. GRPO then normalizes rewards within the group to obtain
\begin{equation}
    A_j =
    \frac{
    r_j - \mathrm{mean}(\{r_1,\ldots,r_G\})
    }{
    \mathrm{std}(\{r_1,\ldots,r_G\})
    } .
\end{equation}
The policy is updated with a clipped objective and a KL penalty toward a reference policy:
\begin{equation}
\begin{aligned}
\mathcal{J}(\theta)
=
\mathbb{E}_{j}\Big[
&\min\big(
\rho_j A_j,\,
\mathrm{clip}(\rho_j,1-\epsilon,1+\epsilon)A_j
\big)
\Big] \\
&-
\beta_{\mathrm{KL}}
\mathbb{D}_{\mathrm{KL}}\!\left(
\pi_\theta \,\|\, \pi_{\mathrm{ref}}
\right),
\end{aligned}
\end{equation}
where
\[
\rho_j =
\frac{
\pi_\theta(o_j\mid\mathbf{x})
}{
\pi_{\theta_{\mathrm{old}}}(o_j\mid\mathbf{x})
}.
\]
Here $\epsilon$ is the clipping range, and the KL penalty keeps the updated policy close to $\pi_{\mathrm{ref}}$. Running this stage without hints gives our GRPO baseline.

\subsection{Progressive Hint Injection}
\label{sec:method:phi}

GRPO improves code-set prediction by optimizing sampled responses with an F1 reward, but some gold codes can remain repeatedly missed by later checkpoints. \method{} turns these false negatives into a training-time curriculum. The model is first evaluated without hints to reveal the codes it currently misses, and later GRPO rounds expose sampled missed codes as in-context hints. The reward target remains the full gold set, not the hint subset.

\paragraph{Missed-code curriculum.}
We maintain a per-sample hint pool $H_i^{(t)}$ for each training example. The round-0 pool is initialized from the SFT model's hint-free predictions:
\begin{equation}
    H_i^{(0)} = Y_i \setminus \hat{Y}^{\mathrm{SFT}}_i .
\end{equation}
At the end of round $t$, we run the current checkpoint on the training set without hints, parse the predicted code set $\hat{Y}_i^{(t)}$, and refresh the pool for the next round:
\begin{equation}
    H_i^{(t+1)} = Y_i \setminus \hat{Y}_i^{(t)} .
\end{equation}
Samples with empty pools are skipped in the next \method{} round, so training concentrates on examples with remaining false negatives.

\paragraph{Stochastic hint injection.}
During \method{} training, each retained prompt either remains hint-free or receives a sampled hint subset $h_i \subseteq H_i^{(t)}$. We sample hints without replacement from the pool. The sampling distribution gives higher priority to codes that are infrequent in the corpus or have low recall under the latest checkpoint, with clipping and temperature smoothing to prevent a few rare codes from dominating training. The hinted prompt states that the listed codes were missed previously but are confirmed applicable, includes their descriptions, and instructs the model to include the hinted codes while adding any other applicable codes supported by the note. Varying the number of injected hints and keeping a fraction of prompts hint-free reduces reliance on the hint list alone and keeps the model exposed to note-only training cases.

\paragraph{Test-time invariance.}
Hints are training-time supervision only. At inference, the model receives only the clinical note and must generate the full code set without hints, matching the hint-free setting used by all generative baselines.
\begin{table*}[t!]
\centering
\setlength{\tabcolsep}{0pt}%
\renewcommand{\arraystretch}{0.95}
\caption{\small\textbf{MIMIC-III ICD-9-CM Results.}
Methods are grouped by paradigm:
\textcolor{plmtag}{\textbf{PLM Baselines}},
\textcolor{llmptag}{\textbf{LLM (Prompting)}}, and
\textcolor{llmttag}{\textbf{LLM (Post-training)}}.
Within each split we report Micro / Macro Recall (R), Precision (P), and F1.
All values are percentages. \textbf{Bold} and \underline{underline} mark the best and second-best result in each column, respectively.}
\label{tab:main_m3}
\vspace{-3pt}
\resizebox{\textwidth}{!}{
\begin{threeparttable}
\begin{tabular}{
   @{\hspace{2pt}}l@{\hspace{6pt}}
   l@{\hspace{8pt}}
   c@{\hspace{4pt}}c@{\hspace{4pt}}c
   @{\hspace{8pt}}
   c@{\hspace{4pt}}c@{\hspace{4pt}}c
   @{\hspace{10pt}}
   c@{\hspace{4pt}}c@{\hspace{4pt}}c
   @{\hspace{8pt}}
   c@{\hspace{4pt}}c@{\hspace{4pt}}c
   @{\hspace{2pt}}
}
\toprule
\multirow{3}{*}{\textbf{Method}} & \multirow{3}{*}{\textbf{Backbone}} &
\multicolumn{6}{c}{\textbf{Top-50}} &
\multicolumn{6}{c}{\textbf{Full}} \\
\cmidrule(lr){3-8} \cmidrule(lr){9-14}
& & \multicolumn{3}{c}{\textbf{Micro}} & \multicolumn{3}{c}{\textbf{Macro}}
  & \multicolumn{3}{c}{\textbf{Micro}} & \multicolumn{3}{c}{\textbf{Macro}} \\
\cmidrule(lr){3-5}\cmidrule(lr){6-8}\cmidrule(lr){9-11}\cmidrule(lr){12-14}
& & Recall & Precision & F1 & Recall & Precision & F1 & Recall & Precision & F1 & Recall & Precision & F1 \\
\midrule

\multicolumn{14}{c}{\textcolor{plmtag}{\ding{182}} \textbf{\textcolor{plmtag}{PLM Baselines}}} \\
CNN          & \llmname{CNN}      & 59.7 & 65.6 & 62.5 & 52.4 & 57.4 & 52.6 & 40.3 & 48.9 & 44.2 & 8.5  & 11.3 & 8.8  \\
GRU          & \llmname{Bi-GRU}   & 36.3 & 56.1 & 44.1 & 30.0 & 44.2 & 33.5 & 42.1 & 49.0 & 45.3 & 9.1  & 12.1 & 9.4  \\
CAML         & \llmname{CNN}      & 45.1 & 69.0 & 54.5 & 38.9 & 56.8 & 44.4 & 48.8 & 53.1 & 50.9 & 17.9 & 20.6 & 18.0 \\
MultiResCNN  & \llmname{ResNet}   & 62.3 & 65.6 & 63.9 & 58.1 & 57.7 & 56.2 & 51.7 & 55.2 & 53.4 & 16.3 & 17.1 & 15.3 \\
LAAT         & \llmname{Bi-LSTM}  & 59.7 & 71.4 & 65.0 & 53.0 & 64.3 & 56.0 & \underline{52.2} & 57.3 & 54.6 & \underline{20.5} & \underline{24.7} & \underline{20.7} \\
PLM-ICD      & \llmname{RoBERTa}  & 68.1 & 68.0 & 68.1 & 63.3 & 63.7 & 63.5 & \textbf{58.4} & 61.0 & \textbf{59.7} & \textbf{29.0} & \textbf{31.4} & \textbf{28.3} \\
\midrule

\multicolumn{14}{c}{\textcolor{llmptag}{\ding{183}} \textbf{\textcolor{llmptag}{LLM (Prompting)}}} \\
\multirow{2}{*}{Zero-shot} & \llmname{Qwen2.5-1.5B} & 0.8 & 29.0 & 1.6 & 0.4 & 4.7 & 0.7 & 0.1 & 15.1 & 0.2 & 0.0 & 0.1 & 0.0 \\
                           & \llmname{Qwen3-4B}     & 5.5 & 52.7 & 10.0 & 2.7 & 11.7 & 4.0 & 1.6 & 24.2 & 3.0 & 0.2 & 0.4 & 0.2 \\
\multirow{2}{*}{Few-shot}  & \llmname{Qwen2.5-1.5B} & 2.2 & 16.1 & 3.9 & 1.4 & 5.6 & 1.9 & 0.0 & 3.9 & 0.0 & 0.0 & 0.0 & 0.0 \\
                           & \llmname{Qwen3-4B}     & 4.6 & 33.4 & 8.0 & 2.8 & 12.1 & 3.9 & 1.1 & 11.8 & 2.0 & 0.1 & 0.4 & 0.1 \\
\multirow{2}{*}{Zero-shot+CoT}      & \llmname{Qwen2.5-1.5B} & 0.8 & 28.0 & 1.5 & 0.3 & 6.9 & 0.6 & 0.0 & 15.1 & 0.1 & 0.0 & 0.1 & 0.0 \\
                           & \llmname{Qwen3-4B}     & 5.8 & 61.5 & 10.6 & 2.8 & 18.0 & 4.2 & 1.9 & 30.8 & 3.5 & 0.2 & 0.6 & 0.2 \\
\multirow{2}{*}{Few-shot + CoT}
& \llmname{Qwen2.5-1.5B}  & 5.1 & 17.8 & 8.0 & 3.3 & 4.6 & 2.8 & 0.6 & 6.9 & 1.1 & 0.0 & 0.1 & 0.0 \\
                           & \llmname{Qwen3-4B}     & 11.0 & 52.8 & 18.2 & 5.9 & 18.2 & 7.8 & 3.2 & 22.9 & 5.7 & 0.3 & 0.8 & 0.3 \\

\midrule

\multicolumn{14}{c}{\textcolor{llmttag}{\ding{184}} \textbf{\textcolor{llmttag}{LLM (Post-training)}}} \\
\multirow{2}{*}{SFT}       & \llmname{Qwen2.5-1.5B} & \underline{73.8} & 65.4 & 69.4 & \underline{68.8} & 64.1 & 65.2 & 30.1 & 45.7 & 36.3 & 9.1 & 14.1 & 10.1 \\
                           & \llmname{Qwen3-4B}     & \textbf{75.2} & 70.1 & 72.6 & \textbf{70.3} & 67.9 & \underline{68.3} & 39.3 & 55.0 & 45.8 & 15.9 & 22.7 & 17.4 \\
\multirow{2}{*}{GRPO}      & \llmname{Qwen2.5-1.5B} & 69.6 & 71.9 & 70.7 & 64.1 & 67.8 & 64.6 & 46.4 & 52.1 & 49.1 & 13.8 & 15.0 & 13.0 \\
                           & \llmname{Qwen3-4B}     & 71.7 & \textbf{75.6} & \textbf{73.6} & 66.5 & \textbf{72.2} & 68.1 & 51.4 & \underline{62.8} & 56.5 & 19.4 & 22.7 & 19.4 \\
\multirow{2}{*}{\textbf{\method{} (Ours)}}
                           & \llmname{Qwen2.5-1.5B} & 72.3 & 68.9 & 70.6 & 67.0 & 66.3 & 65.3 & 45.0 & 55.7 & 49.8 & 15.9 & 15.2 & 14.1 \\
                           & \llmname{Qwen3-4B}     & 73.7 & \underline{73.3} & \underline{73.5} & 68.7 & \underline{70.4} & \textbf{68.6} & 51.4 & \textbf{62.9} & \underline{56.6} & \underline{20.5} & 23.0 & 20.0 \\
\bottomrule
\end{tabular}
\end{threeparttable}
}
\vspace{-4mm}
\end{table*}

\section{Experiments}
\label{sec:exp}

We evaluate the staged post-training progression for generative ICD coding on MIMIC-III~\citep{johnson2016mimic} and MIMIC-IV~\citep{johnson2023mimic}. Our experiments ask three questions. First, can task-specific post-training move generative LLM coders beyond weak prompting-only performance and toward strong discriminative baselines? Second, how much does each stage, SFT, GRPO, and \method{}, contribute to this progression? Third, how does \method{} affect the remaining missed-code cases, including rare codes that prior benchmarks identify as especially challenging?

\subsection{Experimental Setup}
\label{sec:expt:setup}

\paragraph{Datasets.}
We evaluate on MIMIC-III with ICD-9-CM codes and MIMIC-IV with ICD-10-CM codes~\citep{johnson2016mimic,johnson2023mimic}. Both datasets contain discharge summaries from the Beth Israel Deaconess Medical Center. We follow the stratified \textit{clean} splits introduced by \citet{edin2023automated} and report results on both the Top-50 and Full label settings. Our preprocessing preserves punctuation and document structure, which provide useful cues for section boundaries, abbreviations, negation, and clinical lists in autoregressive generation. To ensure a fair comparison, we train or evaluate all discriminative and generative methods on the same preprocessed corpus. Dataset statistics, preprocessing details, and split sizes are provided in Appendix~\ref{app:data}.

\begin{table*}[t!]
\centering
\setlength{\tabcolsep}{0pt}%
\renewcommand{\arraystretch}{0.95}
\caption{\small\textbf{MIMIC-IV ICD-10-CM Results.}
Methods are grouped by paradigm:
\textcolor{plmtag}{\textbf{PLM Baselines}},
\textcolor{llmptag}{\textbf{LLM (Prompting)}}, and
\textcolor{llmttag}{\textbf{LLM (Post-training)}}.
Within each split we report Micro / Macro Recall (R), Precision (P), and F1.
All values are percentages. \textbf{Bold} and \underline{underline} mark the best and second-best result in each column, respectively.}
\label{tab:main_m4}
\vspace{-3pt}
\resizebox{\textwidth}{!}{
\begin{threeparttable}
\begin{tabular}{
   @{\hspace{2pt}}l@{\hspace{6pt}}
   l@{\hspace{8pt}}
   c@{\hspace{4pt}}c@{\hspace{4pt}}c
   @{\hspace{8pt}}
   c@{\hspace{4pt}}c@{\hspace{4pt}}c
   @{\hspace{10pt}}
   c@{\hspace{4pt}}c@{\hspace{4pt}}c
   @{\hspace{8pt}}
   c@{\hspace{4pt}}c@{\hspace{4pt}}c
   @{\hspace{2pt}}
}
\toprule
\multirow{3}{*}{\textbf{Method}} & \multirow{3}{*}{\textbf{Backbone}} &
\multicolumn{6}{c}{\textbf{Top-50}} &
\multicolumn{6}{c}{\textbf{Full}} \\
\cmidrule(lr){3-8} \cmidrule(lr){9-14}
& & \multicolumn{3}{c}{\textbf{Micro}} & \multicolumn{3}{c}{\textbf{Macro}}
  & \multicolumn{3}{c}{\textbf{Micro}} & \multicolumn{3}{c}{\textbf{Macro}} \\
\cmidrule(lr){3-5}\cmidrule(lr){6-8}\cmidrule(lr){9-11}\cmidrule(lr){12-14}
& & Recall & Precision & F1 & Recall & Precision & F1 & Recall & Precision & F1 & Recall & Precision & F1 \\
\midrule

\multicolumn{14}{c}{\textcolor{plmtag}{\ding{182}} \textbf{\textcolor{plmtag}{PLM Baselines}}} \\
CNN          & \llmname{CNN}      & 71.6 & 69.8 & 70.7 & 66.0 & 65.9 & 64.4 & 41.8 & 53.2 & 46.8 & 5.6  & 9.4  & 6.3  \\
GRU          & \llmname{Bi-GRU}   & 71.9 & 70.3 & 71.1 & 66.9 & 65.5 & 65.4 & 43.4 & 54.6 & 48.3 & 8.6  & 13.2 & 9.4  \\
CAML         & \llmname{CNN}      & 70.3 & 68.6 & 69.4 & 64.3 & 64.6 & 63.4 & 51.5 & 58.1 & 54.6 & 14.9 & 18.0 & 15.1 \\
MultiResCNN  & \llmname{ResNet}   & 71.7 & 70.5 & 71.1 & 67.2 & 65.1 & 64.8 & 52.0 & 58.7 & 55.2 & 16.9 & 17.9 & 15.7 \\
LAAT         & \llmname{Bi-LSTM}  & 71.9 & 71.6 & 71.8 & 66.9 & 67.2 & 65.9 & 53.8 & 59.3 & 56.4 & 18.8 & 21.7 & 18.3 \\
PLM-ICD      & \llmname{RoBERTa}  & 73.3 & 73.6 & \underline{73.5} & 68.9 & 69.8 & 68.3 & \textbf{57.0} & 62.4 & \textbf{59.6} & \textbf{24.1} & \textbf{26.8} & \textbf{23.7} \\
\midrule

\multicolumn{14}{c}{\textcolor{llmptag}{\ding{183}} \textbf{\textcolor{llmptag}{LLM (Prompting)}}} \\
\multirow{2}{*}{Zero-shot}
& \llmname{Qwen2.5-1.5B} & 0.1 & 15.0 & 0.3 & 0.1 & 17.0 & 0.2 & 0.1 & 14.4 & 0.1 & 0.0 & 0.6 & 0.0 \\
& \llmname{Qwen3-4B} & 0.4 & 34.3 & 0.9 & 0.4 & 24.4 & 0.8 & 0.3 & 18.9 & 0.6 & 0.1 & 0.7 & 0.1 \\
\multirow{2}{*}{Few-shot}
& \llmname{Qwen2.5-1.5B} & 1.1 & 13.5 & 2.0 & 0.8 & 6.0 & 1.1 & 0.4 & 5.7 & 0.7 & 0.0 & 0.1 & 0.0 \\
& \llmname{Qwen3-4B} & 5.2 & 17.0 & 7.9 & 5.9 & 27.3 & 6.2 & 0.1 & 4.9 & 0.2 & 0.0 & 0.3 & 0.0 \\
\multirow{2}{*}{Zero-shot + CoT}
& \llmname{Qwen2.5-1.5B} & 0.2 & 13.3 & 0.4 & 0.2 & 18.0 & 0.3 & 0.3 & 17.1 & 0.6 & 0.0 & 0.5 & 0.0 \\
& \llmname{Qwen3-4B} & 1.8 & 45.3 & 3.5 & 1.9 & 29.8 & 3.2 & 1.6 & 24.2 & 3.0 & 0.2 & 1.2 & 0.2 \\
\multirow{2}{*}{Few-shot + CoT}
& \llmname{Qwen2.5-1.5B} & 0.5 & 15.2 & 1.0 & 0.3 & 4.5 & 0.6 & 0.2 & 6.3 & 0.3 & 0.0 & 0.1 & 0.0 \\
& \llmname{Qwen3-4B} & 7.3 & 28.6 & 11.7 & 6.9 & 27.4 & 8.5 & 1.5 & 11.1 & 2.6 & 0.2 & 0.9 & 0.2 \\
\midrule

\multicolumn{14}{c}{\textcolor{llmttag}{\ding{184}} \textbf{\textcolor{llmttag}{LLM (Post-training)}}} \\
\multirow{2}{*}{SFT}       & \llmname{Qwen2.5-1.5B} & \textbf{74.3} & 69.6 & 71.9 & \underline{71.7} & 67.6 & 69.2 & 50.3 & 48.8 & 49.5 &	12.8 &	13.1 & 11.8\\
                           & \llmname{Qwen3-4B}     & \textbf{74.3} & 71.7 & 73.0 & \textbf{71.8} & 69.6 & \textbf{70.3} & 46.8 & 58.5 & 52.0 & 19.3 & 24.2 & 20.0 \\
\multirow{2}{*}{GRPO}      & \llmname{Qwen2.5-1.5B} & 71.1 & \underline{74.1} & 72.6 & 68.5 & 71.0 & 68.8 & 49.1 & 58.5 & 53.3 & 10.7 & 13.8 & 10.8 \\
                           & \llmname{Qwen3-4B}     & 71.5 & \textbf{75.9} & \textbf{73.6} & 68.8 & \textbf{73.1} & 69.8 & 54.2 & \textbf{63.7} & 58.6 & 20.2 & 24.9 & 20.6 \\
\multirow{2}{*}{\textbf{\method{} (Ours)}}
                           & \llmname{Qwen2.5-1.5B} & \underline{73.8} & 70.8 & 72.3 & 71.1 & 68.6 & 69.3 & 51.0 & 57.0 & 53.8 & 13.2 & 13.9 & 12.3 \\
                           & \llmname{Qwen3-4B}     & 73.7 & 73.2 & 73.4 & 70.8 & \underline{71.1} & \underline{70.1} & \underline{55.0} & \underline{63.0} & \underline{58.7} & \underline{21.0} & \underline{25.0} & \underline{21.2} \\
\bottomrule
\end{tabular}
\end{threeparttable}
}
\vspace{-4mm}
\end{table*}

\paragraph{Baselines.}
We compare two families of ICD coders. \textbf{Discriminative methods} include CNN, GRU, and CAML~\citep{mullenbach2018explainable}, MultiResCNN~\citep{li2020icd}, LAAT~\citep{vu2020label}, and PLM-ICD~\citep{huang2022plm}. \textbf{Generative methods} use \llmname{Qwen2.5-1.5B} and \llmname{Qwen3-4B} as shared backbones and cover prompting variants, supervised fine-tuning, GRPO~\citep{shao2024deepseekmath}, and \method{}. Prompting variants include zero-shot, few-shot, and chain-of-thought prompting~\citep{wei2022chain}. All generative variants are evaluated with the same parser, output format, and metric set to ensure a controlled comparison. Baseline configurations and prompt templates are in Appendix~\ref{app:baselines}.

\paragraph{Evaluation Metrics.}
We report precision, recall, and F1 at both the micro and macro levels, with all values shown as percentages. Micro scores are computed by pooling true positives, false positives, and false negatives over the test set, while macro scores are computed as the arithmetic mean of per-class scores following \citet{edin2023automated}. For discriminative models, we tune a single decision threshold on the validation set to maximize micro-F1. Generative models do not use a decision threshold. Full metric definitions are provided in Appendix~\ref{app:metrics}.

\paragraph{Implementation Details.}
For discriminative baselines, we adopt the tuned hyperparameters of \citet{edin2023automated} and retrain each model on our preprocessed corpus. For generative methods, we use LoRA for SFT and GRPO-based RL training with vLLM-backed rollouts. All post-training variants share the same backbone, parser, and reward implementation. Hardware, hyperparameters, and training-time details are provided in Appendix~\ref{app:impl}.

\subsection{Main Results}

Tables~\ref{tab:main_m3} and~\ref{tab:main_m4} show a clear shift in how generative LLM coders should be interpreted. Under prompting alone, both backbones perform far below discriminative coders, especially in the Full label setting, which is consistent with prior reports that LLMs are weak medical coders. However, this comparison mainly reflects the limitation of prompting-only evaluation rather than the capability of task-adapted generative coding. After task-specific post-training, the conclusion changes substantially. SFT provides the main capability jump, GRPO improves code-set optimization in the larger label space, and \method{} acts as a focused refinement stage for remaining missed codes. Overall, the results suggest that generative LLMs are not inherently poor ICD coders. Rather, prompting-only evaluation conflates the limits of prompting with the potential of task-adapted generative coding.

\paragraph{SFT provides the main capability jump.}
The largest improvement comes from supervised fine-tuning. Prompted LLMs often fail not only because they lack ICD-specific knowledge, but also because they do not reliably follow the required output schema, generate invalid codes, or place codes outside the parseable \texttt{<code>...</code>} span. SFT addresses this first bottleneck by training the model on note-code demonstrations, teaching both a parseable output format and the empirical ICD code distribution. For example, on MIMIC-III Top-50, \llmname{Qwen3-4B} improves from the best prompting micro-F1 of 18.2 to 72.6 after SFT, already surpassing PLM-ICD. Similar jumps appear on MIMIC-IV and in the Full setting, where SFT brings the model from near-unusable prompting performance into a strong working range. This shows that the main barrier for prompted LLMs is not the generative formulation itself, but the lack of task-specific adaptation.

\paragraph{GRPO improves code-set prediction beyond SFT.}
After SFT establishes a usable generative coder, GRPO provides the main metric-oriented refinement. Its effect is most visible in the Full setting, where the model must balance precision and recall over a much larger ICD taxonomy. Across both MIMIC-III and MIMIC-IV, GRPO consistently improves Full-setting micro-F1 over SFT, with especially large gains on MIMIC-III. In contrast, Top-50 results are already strong after SFT, so GRPO mainly adjusts the precision--recall tradeoff rather than producing another large jump. This pattern suggests that reward-based post-training is most useful once the model has learned the coding format and code prior, but still needs to optimize complete code-set quality in a large label space.

\paragraph{\method{} refines the remaining missed-code cases.}
Unlike SFT and GRPO, this stage is designed for focused refinement rather than broad capability acquisition. It starts from the SFT-initialized policy and repeatedly trains on cases where the current checkpoint still misses gold codes. Samples whose hint pools become empty are skipped in later \method{} rounds, which makes training faster and concentrates updates on unresolved cases. This design can reduce repeated exposure to easy or already-solved cases that contribute heavily to micro-F1. As a result, \method{} is often close to GRPO on headline micro-F1 rather than uniformly higher. Its benefit is more visible in the Full setting and in macro-level performance, where remaining missed and lower-frequency codes matter more. This makes \method{} a targeted complement to GRPO: GRPO improves overall code-set quality, while \method{} redirects later training toward the codes that the current policy still fails to recover.

\paragraph{Post-trained LLMs narrow the gap to PLM-ICD.}
PLM-ICD remains a strong reference point, especially in the Full label setting. Still, post-training substantially changes the comparison. In Top-50 settings, post-trained LLMs reach or exceed PLM-ICD on MIMIC-III and match or slightly exceed it on MIMIC-IV. In Full settings, they do not uniformly surpass PLM-ICD, but GRPO and \method{} close much of the gap while retaining the generative formulation. Model size helps, especially after SFT and GRPO in the Full setting, but it is not the main driver: prompted \llmname{Qwen3-4B} remains far below post-trained \llmname{Qwen2.5-1.5B}. Thus, the main conclusion is not that generative LLMs replace discriminative coders outright, but that task-specific post-training makes them competitive in settings where prompting alone fails.
\section{Future Directions}
\label{sec:future}

Our results suggest that future generative ICD coders should focus on the Full label setting, where the key bottleneck is rare-code recall. Discriminative PLM coders score every label with an explicit classifier head, giving them a natural mechanism for label coverage and threshold-based recall control. Generative LLM coders are different: they must actively recall and emit the correct codes from a large taxonomy, so rare codes can be omitted even when the note contains supporting evidence. \method{} shows that training-time exposure to previously missed codes is a promising way to improve this behavior, but stronger mechanisms are needed to fully close the Full-setting gap.

\paragraph{Agentic RL with retrieval.}
A natural direction is to let the model learn when to consult external coding resources before producing the final code set. During RL, the policy could decide whether to retrieve ICD descriptions, hierarchy neighbors, or similar coded cases, then use this evidence to improve recall. This is especially relevant for rare codes, where the model may not have enough parametric knowledge to generate the correct code without assistance.

\paragraph{Hybrid PLM--LLM coding.}
Another direction is to combine the label coverage of discriminative coders with the flexibility of generative LLMs. A PLM classifier can provide a high-recall candidate set over the full taxonomy, while the LLM verifies, completes, or explains the final prediction. Such systems may be especially useful in the Full setting, where exhaustive label scoring remains a major advantage of discriminative models.

\paragraph{Recall-oriented rewards and decoding.}
Our GRPO reward uses sample-level F1, which balances precision and recall. Future work could explore rewards and decoding strategies that explicitly emphasize rare-code recall, such as $F_\beta$ rewards, class-balanced rewards, bonuses for repeatedly missed codes, or recall-calibrated stopping rules. The key challenge is to improve low-frequency code recovery without encouraging broad overprediction and precision collapse.

\section{Conclusion}

We presented a controlled empirical study of task-specific post-training for generative ICD coding. By comparing prompting, SFT, and RL-based post-training under the same protocol, this study provides the first evaluation of whether RL-based post-training can improve generative LLM coders for ICD code-set prediction. Our results show that prompting-only evaluation substantially underestimates the potential of LLM coders: SFT provides the main capability jump, GRPO improves code-set prediction in large label spaces, and \method{} offers a targeted curriculum for remaining missed-code cases. While strong discriminative PLM coders remain highly competitive, task-specific post-training narrows the gap and makes generative LLMs viable ICD coders under controlled evaluation. More broadly, these findings shift the central question from whether LLMs can code from prompting alone to how post-training, retrieval, and reward mechanisms should be designed to improve reliable recall over the full ICD taxonomy.

\section{Limitations}

Our evaluation is bounded in several ways. First, we use MIMIC-III and MIMIC-IV, which are standard ICD coding benchmarks but come from a limited clinical data ecosystem. The results may not fully reflect documentation styles, coding practices, or label distributions across institutions. Second, due to computational constraints, our generative experiments cover two relatively small open-source Qwen backbones. Larger LLMs, other model families, and closed-source models may show different scaling behavior under the same post-training pipeline. Finally, the Full label setting remains challenging, especially for rare-code recall, and post-training does not fully close the gap to PLM-ICD.
\bibliography{main}

@article{johnson2016mimic,
  title={MIMIC-III, a freely accessible critical care database},
  author={Johnson, Alistair EW and Pollard, Tom J and Shen, Lu and Lehman, Li-wei H and Feng, Mengling and Ghassemi, Mohammad and Moody, Benjamin and Szolovits, Peter and Anthony Celi, Leo and Mark, Roger G},
  journal={Scientific data},
  volume={3},
  number={1},
  pages={1--9},
  year={2016},
  publisher={Nature Publishing Group}
}

@article{johnson2023mimic,
  title={MIMIC-IV, a freely accessible electronic health record dataset},
  author={Johnson, Alistair EW and Bulgarelli, Lucas and Shen, Lu and Gayles, Alvin and Shammout, Ayad and Horng, Steven and Pollard, Tom J and Hao, Sicheng and Moody, Benjamin and Gow, Brian and others},
  journal={Scientific data},
  volume={10},
  number={1},
  pages={1},
  year={2023},
  publisher={Nature Publishing Group UK London}
}

@inproceedings{mullenbach2018explainable,
  title={Explainable Prediction of Medical Codes from Clinical Text},
  author={Mullenbach, James and Wiegreffe, Sarah and Duke, Jon and Sun, Jimeng and Eisenstein, Jacob},
  booktitle={Proceedings of the 2018 Conference of the North American Chapter of the Association for Computational Linguistics: Human Language Technologies, Volume 1 (Long Papers)},
  pages={1101--1111},
  year={2018},
  address={New Orleans, Louisiana},
  publisher={Association for Computational Linguistics},
  url={https://aclanthology.org/N18-1100/},
  doi={10.18653/v1/N18-1100}
}

@inproceedings{vu2020label,
  title={A Label Attention Model for ICD Coding from Clinical Text},
  author={Vu, Thanh and Nguyen, Dat Quoc and Nguyen, Anthony},
  booktitle={Proceedings of the Twenty-Ninth International Joint Conference on Artificial Intelligence},
  pages={3335--3341},
  year={2020},
  doi={10.24963/ijcai.2020/461}
}

@inproceedings{huang2022plm,
  title={{PLM}-{ICD}: Automatic {ICD} Coding with Pretrained Language Models},
  author={Huang, Chao-Wei and Tsai, Shang-Chi and Chen, Yun-Nung},
  booktitle={Proceedings of the 4th Clinical Natural Language Processing Workshop},
  pages={10--20},
  year={2022},
  address={Seattle, WA},
  publisher={Association for Computational Linguistics},
  url={https://aclanthology.org/2022.clinicalnlp-1.2/},
  doi={10.18653/v1/2022.clinicalnlp-1.2}
}

@article{boyle2023automated,
  title={Automated clinical coding using off-the-shelf large language models},
  author={Boyle, Joseph S and Kascenas, Antanas and Lok, Pat and Liakata, Maria and O'Neil, Alison Q},
  journal={arXiv preprint arXiv:2310.06552},
  year={2023}
}

@article{kwan2024large,
  title={Large language models are good medical coders, if provided with tools},
  author={Kwan, Keith},
  journal={arXiv preprint arXiv:2407.12849},
  year={2024}
}

@article{teng2022review,
  title={A review on deep neural networks for ICD coding},
  author={Teng, Fei and Liu, Yiming and Li, Tianrui and Zhang, Yi and Li, Shuangqing and Zhao, Yue},
  journal={IEEE Transactions on Knowledge and Data Engineering},
  volume={35},
  number={5},
  pages={4357--4375},
  year={2022},
  publisher={IEEE}
}

@article{ji2024unified,
  title={A unified review of deep learning for automated medical coding},
  author={Ji, Shaoxiong and Li, Xiaobo and Sun, Wei and Dong, Hang and Taalas, Ara and Zhang, Yijia and Wu, Honghan and Pitk{\"a}nen, Esa and Marttinen, Pekka},
  journal={ACM Computing Surveys},
  volume={56},
  number={12},
  pages={1--41},
  year={2024},
  publisher={ACM New York, NY}
}

@article{hou2025enhancing,
  title={Enhancing medical coding efficiency through domain-specific fine-tuned large language models},
  author={Hou, Zhen and Liu, Hao and Bian, Jiang and He, Xing and Zhuang, Yan},
  journal={npj Health Systems},
  volume={2},
  number={1},
  pages={14},
  year={2025},
  publisher={Nature Publishing Group UK London}
}

@inproceedings{zhang2020bert,
  title={BERT-XML: Large scale automated ICD coding using BERT pretraining},
  author={Zhang, Zachariah and Liu, Jingshu and Razavian, Narges},
  booktitle={Proceedings of the 3rd Clinical Natural Language Processing Workshop},
  pages={24--34},
  year={2020}
}

@inproceedings{li2020icd,
  title={ICD coding from clinical text using multi-filter residual convolutional neural network},
  author={Li, Fei and Yu, Hong},
  booktitle={proceedings of the AAAI conference on artificial intelligence},
  volume={34},
  pages={8180--8187},
  year={2020}
}

@inproceedings{liu2021effective,
  title={Effective convolutional attention network for multi-label clinical document classification},
  author={Liu, Yang and Cheng, Hua and Klopfer, Russell and Gormley, Matthew R and Schaaf, Thomas},
  booktitle={Proceedings of the 2021 Conference on Empirical Methods in Natural Language Processing},
  pages={5941--5953},
  year={2021}
}

@inproceedings{tsai2021modeling,
  title={Modeling diagnostic label correlation for automatic ICD coding},
  author={Tsai, Shang-Chi and Huang, Chao-Wei and Chen, Yun-Nung},
  booktitle={Proceedings of the 2021 conference of the North American chapter of the association for computational linguistics: human language technologies},
  pages={4043--4052},
  year={2021}
}

@article{wei2022chain,
  title={Chain-of-thought prompting elicits reasoning in large language models},
  author={Wei, Jason and Wang, Xuezhi and Schuurmans, Dale and Bosma, Maarten and Xia, Fei and Chi, Ed and Le, Quoc V and Zhou, Denny and others},
  journal={Advances in neural information processing systems},
  volume={35},
  pages={24824--24837},
  year={2022}
}

@article{shao2024deepseekmath,
  title={Deepseekmath: Pushing the limits of mathematical reasoning in open language models},
  author={Shao, Zhihong and Wang, Peiyi and Zhu, Qihao and Xu, Runxin and Song, Junxiao and Bi, Xiao and Zhang, Haowei and Zhang, Mingchuan and Li, YK and Wu, Yang and others},
  journal={arXiv preprint arXiv:2402.03300},
  year={2024}
}

@article{soroush2024large,
  title={Large language models are poor medical coders—benchmarking of medical code querying},
  author={Soroush, Ali and Glicksberg, Benjamin S and Zimlichman, Eyal and Barash, Yiftach and Freeman, Robert and Charney, Alexander W and Nadkarni, Girish N and Klang, Eyal},
  journal={Nejm Ai},
  volume={1},
  number={5},
  pages={AIdbp2300040},
  year={2024},
  publisher={Massachusetts Medical Society}
}

@inproceedings{li2026evaluation,
  title={Evaluation and LLM-Guided Learning of ICD Coding Rationales},
  author={Li, Mingyang and Schlegel, Viktor and Mu, Tingting and Oyewusi, Wuraola and Kang, Kai and Nenadic, Goran},
  booktitle={Proceedings of the 19th Conference of the European Chapter of the Association for Computational Linguistics (Volume 1: Long Papers)},
  pages={4969--5003},
  year={2026}
}

@inproceedings{baksi2025medcoder,
  title={MedCodER: A generative AI assistant for medical coding},
  author={Baksi, Krishanu Das and Soba, Elijah and Higgins, John J and Saini, Ravi and Wood, Jaden and Cook, Jane and Scott, Jack I and Pudota, Nirmala and Weninger, Tim and Bowen, Edward and others},
  booktitle={Proceedings of the 2025 Conference of the Nations of the Americas Chapter of the Association for Computational Linguistics: Human Language Technologies (Volume 3: Industry Track)},
  pages={449--459},
  year={2025}
}

@inproceedings{edin2023automated,
  title={Automated medical coding on mimic-iii and mimic-iv: A critical review and replicability study},
  author={Edin, Joakim and Junge, Alexander and Havtorn, Jakob D and Borgholt, Lasse and Maistro, Maria and Ruotsalo, Tuukka and Maal{\o}e, Lars},
  booktitle={Proceedings of the 46th international ACM SIGIR conference on research and development in information retrieval},
  pages={2572--2582},
  year={2023}
}

@article{hu2022lora,
  title={Lora: Low-rank adaptation of large language models.},
  author={Hu, Edward J and Shen, Yelong and Wallis, Phillip and Allen-Zhu, Zeyuan and Li, Yuanzhi and Wang, Shean and Wang, Liang and Chen, Weizhu and others},
  journal={Iclr},
  volume={1},
  number={2},
  pages={3},
  year={2022}
}

@article{lu2025himarl,
  title={Enhancing ICD code assignment with hierarchical multi-agent collaboration and reinforcement learning},
  author={Lu, Pengli and Yang, Xue and Xue, Jingjin and Gao, Fentang},
  journal={Biomedical Signal Processing and Control},
  volume={110},
  pages={109094},
  year={2025},
  publisher={Elsevier},
  doi={10.1016/j.bspc.2025.109094}
}

@article{ouyang2022training,
  title={Training language models to follow instructions with human feedback},
  author={Ouyang, Long and Wu, Jeffrey and Jiang, Xu and Almeida, Diogo and Wainwright, Carroll and Mishkin, Pamela and Zhang, Chong and Agarwal, Sandhini and Slama, Katarina and Ray, Alex and others},
  journal={Advances in neural information processing systems},
  volume={35},
  pages={27730--27744},
  year={2022}
}

@article{wei2021finetuned,
  title={Finetuned language models are zero-shot learners},
  author={Wei, Jason and Bosma, Maarten and Zhao, Vincent Y and Guu, Kelvin and Yu, Adams Wei and Lester, Brian and Du, Nan and Dai, Andrew M and Le, Quoc V},
  journal={arXiv preprint arXiv:2109.01652},
  year={2021}
}

@article{ranzato2015sequence,
  title={Sequence level training with recurrent neural networks},
  author={Ranzato, Marc'Aurelio and Chopra, Sumit and Auli, Michael and Zaremba, Wojciech},
  journal={arXiv preprint arXiv:1511.06732},
  year={2015}
}

@article{schulman2017proximal,
  title={Proximal policy optimization algorithms},
  author={Schulman, John and Wolski, Filip and Dhariwal, Prafulla and Radford, Alec and Klimov, Oleg},
  journal={arXiv preprint arXiv:1707.06347},
  year={2017}
}

@inproceedings{alsentzer2019publicly,
  title={Publicly available clinical BERT embeddings},
  author={Alsentzer, Emily and Murphy, John and Boag, William and Weng, Wei-Hung and Jindi, Di and Naumann, Tristan and McDermott, Matthew},
  booktitle={Proceedings of the 2nd clinical natural language processing workshop},
  pages={72--78},
  year={2019}
}

@article{gu2021domain,
  title={Domain-specific language model pretraining for biomedical natural language processing},
  author={Gu, Yu and Tinn, Robert and Cheng, Hao and Lucas, Michael and Usuyama, Naoto and Liu, Xiaodong and Naumann, Tristan and Gao, Jianfeng and Poon, Hoifung},
  journal={ACM Transactions on Computing for Healthcare (HEALTH)},
  volume={3},
  number={1},
  pages={1--23},
  year={2021},
  publisher={ACM New York, NY}
}

\clearpage
\newpage

\appendix
\section{Dataset Statistics and Preprocessing}
\label{app:data}

\paragraph{Sources.}
All MIMIC-based experiments use credentialed-access clinical data released through PhysioNet. For MIMIC-III~\citep{johnson2016mimic}, we use discharge summaries paired with ICD-9-CM diagnosis and procedure codes. For MIMIC-IV~\citep{johnson2023mimic}, we use MIMIC-IV and MIMIC-IV-Note discharge summaries paired with ICD-10-CM diagnosis and ICD-10-PCS procedure codes. Both datasets require completion of the appropriate PhysioNet credentialing and data-use agreements.

\paragraph{Preprocessing and splits.}
We follow the preprocessing pipeline of \citet{edin2023automated} to convert the raw note and code-assignment tables into sectioned-note datasets. Each processed example contains a discharge summary, its deduplicated target code set, a split identifier, and a note identifier. We use the corresponding clean splits from \citet{edin2023automated}, which remove rare codes with fewer than ten training occurrences and use multi-label stratified sampling so that evaluated codes appear in both training and test sets. Unlike the original preprocessing, which lowercases text and strips non-alphanumeric characters, we preserve common punctuation while still lowercasing the note text. Punctuation, section boundaries, abbreviations, negation cues, and list structure provide useful signals for autoregressive generation. To keep comparisons fair, all discriminative and generative methods are trained or evaluated on the same punctuation-preserved inputs.

\paragraph{Top-50 vs. Full.}
For both MIMIC-III and MIMIC-IV, we construct two label-set variants. The \textbf{Full} variant retains all diagnosis and procedure codes that remain after preprocessing. The \textbf{Top-50} variant restricts the label space to the 50 most frequent codes in the corresponding training set, following the standard top-code evaluation protocol used in prior medical-coding work. Examples with no labels after Top-50 filtering are excluded from the corresponding Top-50 split.

\paragraph{Instruction-format conversion.}
After preprocessing, each dataset is converted into a ShareGPT-style instruction-tuning format. The user turn contains the discharge summary and an instruction to output all applicable diagnosis and procedure codes. The assistant turn contains the gold code set enclosed in a structured \texttt{<code>...</code>} block. The same conversion is applied to MIMIC-III and MIMIC-IV, yielding Full and Top-50 variants for supervised fine-tuning, prompting evaluation, and reinforcement-learning experiments. Because the processed examples contain restricted clinical text, we do not redistribute the note-level data. Reproducing the datasets requires authorized PhysioNet access and rerunning the preprocessing pipeline.

\section{Models}
\label{app:models}

We use two instruction-tuned Qwen backbones for all generative experiments: \llmname{Qwen2.5-1.5B} and \llmname{Qwen3-4B}. \llmname{Qwen2.5-1.5B} provides a lightweight setting for studying whether post-training can improve a small generative ICD coder, while \llmname{Qwen3-4B} provides a stronger backbone that remains practical for SFT and GRPO-based training. Both models are evaluated under the same prompting, SFT, GRPO, and \method{} settings.

\section{Baseline Configurations}
\label{app:baselines}

\paragraph{Discriminative baselines.}
CNN, GRU, and CAML~\citep{mullenbach2018explainable} share a Word2Vec embedding layer trained on MIMIC notes. CNN and GRU pool the encoder output before classification. CAML applies the label-aware attention from its original paper. MultiResCNN~\citep{li2020icd} replaces the single-filter convolution with a multi-scale residual variant. LAAT~\citep{vu2020label} uses a Bi-LSTM encoder with a label attention layer. PLM-ICD~\citep{huang2022plm} uses a RoBERTa encoder pre-trained on PubMed and MIMIC, with the same label attention as LAAT. All hyperparameters follow \citet{edin2023automated} Table~3.

\paragraph{Generative baselines.}
We use \llmname{Qwen2.5-1.5B} and \llmname{Qwen3-4B} as shared backbones for all generative methods. The prompting baselines include zero-shot, few-shot, zero-shot chain-of-thought, and few-shot chain-of-thought prompting~\citep{wei2022chain}. The training-based generative baselines follow the staged post-training pipeline studied in the main paper: \textit{SFT} fine-tunes the backbone on note-to-code demonstrations, \textit{GRPO} further optimizes the SFT policy with a sample-level F1 reward, and \method{} extends GRPO with progressive training-time hints from missed codes. All generative methods use the same output format, parser, and evaluation metrics.

\definecolor{promptblue}{HTML}{0B3C5D}
\definecolor{promptborder}{HTML}{2E6F82}
\definecolor{promptbg}{HTML}{F7FBFC}
\newcommand{\promptplaceholder}[1]{\textnormal{\textit{\{#1\}}}}
\newcommand{\promptline}[1]{#1\par}
\newcommand{\PromptBox}[2]{%
  \begin{tcolorbox}[
    enhanced,
    breakable,
    colback=promptbg,
    colframe=promptborder,
    colbacktitle=promptblue,
    coltitle=white,
    title={#1},
    fonttitle=\bfseries,
    fontupper=\scriptsize\ttfamily,
    boxrule=0.6pt,
    arc=2.4pt,
    outer arc=2.4pt,
    boxsep=0pt,
    left=0.75em,
    right=0.75em,
    top=0.45em,
    bottom=0.45em,
    toptitle=0.18em,
    bottomtitle=0.18em,
    lefttitle=0.65em,
    before skip=0.4em,
    after skip=0.6em,
    before upper={\setlength{\parindent}{0pt}\setlength{\parskip}{0.12em}},
  ]
  #2
  \end{tcolorbox}%
}

\section{Prompt Templates}
\label{app:prompt_templates}

We summarize the prompt templates used for the prompting baselines below. The placeholders \texttt{\{CORPUS\}}, \texttt{\{CODE\_SYSTEM\}}, and \texttt{\{LABEL\_SCOPE\}} are filled according to the dataset and label-set setting: MIMIC-III or MIMIC-IV, ICD-9-CM or ICD-10-CM/PCS, and Top-50 or Full. Importantly, the prompt does not include an explicit candidate code list. The Top-50 and Full settings determine only the dataset construction and evaluation label space. During evaluation, we parse and score only the final codes enclosed in \texttt{<code>...</code>}.

\paragraph{Zero-shot.}
The zero-shot setting contains no clinical demonstrations. The short code block
shown in the prompt is a formatting example only and is not an allowed-code list.

\PromptBox{Zero-shot Prompt}{
\promptline{You are a clinical coding specialist assigning \promptplaceholder{CODE\_SYSTEM} codes from a discharge summary.}
\promptline{Use \promptplaceholder{LABEL\_SCOPE}.}
\promptline{Output only the final \promptplaceholder{CODE\_SYSTEM} codes in <code>...</code>. Do not include any other text.}
\promptline{}
\promptline{Formatting example only, not a clinical example and not codes to copy:}
\promptline{<code>\promptplaceholder{FORMAT\_EXAMPLE\_CODES}</code>}
\promptline{}
\promptline{Now code the following discharge summary.}
\promptline{Discharge Summary:}
\promptline{\promptplaceholder{DISCHARGE\_SUMMARY}}
\promptline{}
\promptline{Answer:}
}

\paragraph{Few-shot.}
The few-shot setting prepends three retrieved training examples, each consisting
of a discharge summary and its gold code block. Retrieval is based on note
similarity.

\PromptBox{Few-shot Prompt}{
\promptline{You are a clinical coding specialist assigning \promptplaceholder{CODE\_SYSTEM} codes from a discharge summary.}
\promptline{Use \promptplaceholder{LABEL\_SCOPE}.}
\promptline{Output only the final \promptplaceholder{CODE\_SYSTEM} codes in <code>...</code>. Do not include any other text.}
\promptline{}
\promptline{Examples:}
\promptline{}
\promptline{Example 1}
\promptline{Discharge Summary:}
\promptline{\promptplaceholder{SHOT\_NOTE\_1}}
\promptline{}
\promptline{Answer:}
\promptline{<code>\promptplaceholder{SHOT\_CODES\_1}</code>}
\promptline{}
\promptline{Example 2}
\promptline{Discharge Summary:}
\promptline{\promptplaceholder{SHOT\_NOTE\_2}}
\promptline{}
\promptline{Answer:}
\promptline{<code>\promptplaceholder{SHOT\_CODES\_2}</code>}
\promptline{}
\promptline{Example 3}
\promptline{Discharge Summary:}
\promptline{\promptplaceholder{SHOT\_NOTE\_3}}
\promptline{}
\promptline{Answer:}
\promptline{<code>\promptplaceholder{SHOT\_CODES\_3}</code>}
\promptline{}
\promptline{Now code the following discharge summary.}
\promptline{Discharge Summary:}
\promptline{\promptplaceholder{DISCHARGE\_SUMMARY}}
\promptline{}
\promptline{Answer:}
}

\paragraph{Zero-shot + CoT.}
The zero-shot CoT setting asks the model to write a brief structured rationale
inside \texttt{<think>...</think>} before giving the final code block. Only the
codes inside \texttt{<code>...</code>} are scored.

\PromptBox{Zero-shot + CoT Prompt}{
\promptline{You are a clinical coding specialist assigning \promptplaceholder{CODE\_SYSTEM} codes from a discharge summary.}
\promptline{Use \promptplaceholder{LABEL\_SCOPE}.}
\promptline{First write a brief structured rationale in <think>...</think>, then output the final \promptplaceholder{CODE\_SYSTEM} codes in <code>...</code>.}
\promptline{Do not put codes outside the <code> block.}
\promptline{}
\promptline{Formatting example only, not a clinical example and not codes to copy:}
\promptline{<think>Briefly identify the documented diagnoses and procedures, then map them to \promptplaceholder{CODE\_SYSTEM} codes.</think>}
\promptline{<code>\promptplaceholder{FORMAT\_EXAMPLE\_CODES}</code>}
\promptline{}
\promptline{Now code the following discharge summary.}
\promptline{Discharge Summary:}
\promptline{\promptplaceholder{DISCHARGE\_SUMMARY}}
\promptline{}
\promptline{Answer:}
}

\paragraph{Few-shot + CoT.}
The few-shot CoT setting combines the same three retrieved demonstrations with
the structured rationale format. As above, evaluation ignores the rationale and
scores only the final code block.

\PromptBox{Few-shot + CoT Prompt}{
\promptline{You are a clinical coding specialist assigning \promptplaceholder{CODE\_SYSTEM} codes from a discharge summary.}
\promptline{Use \promptplaceholder{LABEL\_SCOPE}.}
\promptline{First write a brief structured rationale in <think>...</think>, then output the final \promptplaceholder{CODE\_SYSTEM} codes in <code>...</code>.}
\promptline{Do not put codes outside the <code> block.}
\promptline{}
\promptline{Examples:}
\promptline{}
\promptline{Example 1}
\promptline{Discharge Summary:}
\promptline{\promptplaceholder{SHOT\_NOTE\_1}}
\promptline{}
\promptline{Answer:}
\promptline{<think>Review the discharge summary for documented diagnoses and procedures, then map the supported findings to \promptplaceholder{CODE\_SYSTEM} codes.</think>}
\promptline{<code>\promptplaceholder{SHOT\_CODES\_1}</code>}
\promptline{}
\promptline{Example 2}
\promptline{Discharge Summary:}
\promptline{\promptplaceholder{SHOT\_NOTE\_2}}
\promptline{}
\promptline{Answer:}
\promptline{<think>Review the discharge summary for documented diagnoses and procedures, then map the supported findings to \promptplaceholder{CODE\_SYSTEM} codes.</think>}
\promptline{<code>\promptplaceholder{SHOT\_CODES\_2}</code>}
\promptline{}
\promptline{Example 3}
\promptline{Discharge Summary:}
\promptline{\promptplaceholder{SHOT\_NOTE\_3}}
\promptline{}
\promptline{Answer:}
\promptline{<think>Review the discharge summary for documented diagnoses and procedures, then map the supported findings to \promptplaceholder{CODE\_SYSTEM} codes.</think>}
\promptline{<code>\promptplaceholder{SHOT\_CODES\_3}</code>}
\promptline{}
\promptline{Now code the following discharge summary.}
\promptline{Discharge Summary:}
\promptline{\promptplaceholder{DISCHARGE\_SUMMARY}}
\promptline{}
\promptline{Answer:}
}

\section{Training Details}
\label{app:training}

\paragraph{SFT.}
We fine-tune with LoRA~\citep{hu2022lora} of rank 8, learning rate $1{\times}10^{-5}$ with cosine schedule, and 3 epochs. The optimizer is AdamW. Each pair is formatted as a single ShareGPT-style user turn followed by the gold code span as the assistant turn. The user segment is masked out of the loss, so the cross-entropy objective is computed only over the assistant response tokens.

\paragraph{GRPO.}
We sample $G=8$ trajectories per prompt and form within-group advantages from standardized sample-level F1 rewards. The actor is updated with the clipped GRPO objective and a KL penalty to the reference policy. The KL coefficient is $1{\times}10^{-3}$ throughout. The maximum response length is 196 tokens, sufficient for any plausible code list. Rollouts use vLLM with top-$p$ sampling.

\paragraph{Reward implementation.}
For each generated response, we extract the predicted code set $\hat{Y}$ from the final \texttt{<code>...</code>} span, normalize and deduplicate the codes, and compare it with the gold set $Y$. The scalar reward passed to GRPO is the sample-level F1 between $\hat{Y}$ and $Y$. Corpus-level micro-F1 and macro-F1 are computed only for logging and evaluation, not used as rollout rewards.

\paragraph{\method{} hint training.}
We run \method{} for 3 to 5 rounds, with one GRPO epoch per round. The round-0 hint pool is initialized from the SFT model's hint-free predictions:
\[
H_i^{(0)} = Y_i \setminus \hat{Y}^{\mathrm{SFT}}_i .
\]
At each subsequent round, we run greedy inference with the previous-round checkpoint on the training set without hints, parse $\hat{Y}_i^{(t-1)}$, and refresh
\[
H_i^{(t)} = Y_i \setminus \hat{Y}_i^{(t-1)} .
\]
Samples whose hint pool becomes empty are filtered out of the next round. At each training step, a hint is injected with probability $0.5$. When injecting, we sample between 1 and 5 missed codes without replacement from $H_i^{(t)}$ using the clipped and temperature-smoothed code weights described below.

\paragraph{Hint sampling.}
For each missed code $c$, we assign a rare-and-hard priority based on its corpus frequency and latest-round recall:
\[
a_c =
\frac{1}{\sqrt{\mathrm{freq}_c}}
\cdot
\frac{1}{\max(\mathrm{recall}_c,0.05)} .
\]
We clip this priority at three times the median priority and sample hints without replacement from the temperature-smoothed distribution
\[
p(c\mid H_i^{(t)})
=
\frac{\tilde{w}_c^{1/\tau}}
{\sum_{c'\in H_i^{(t)}} \tilde{w}_{c'}^{1/\tau}},
\]
where $\tilde{w}_c$ is the clipped priority and $\tau=3.0$. This preserves a bias toward infrequent and low-recall codes while preventing a few rare codes from dominating training.

\paragraph{Hint prompt.}
When hints are injected, we insert a training hint before the output-format instruction. The hint lists the previously missed codes, includes their ICD descriptions, and gives a mini example showing that the final answer should include both the hinted codes and any additional codes supported by the note. The template is:

\begin{quote}
\small
\texttt{Training hint:}\\
\texttt{The following ICD-\{VERSION\} codes were missed previously, but they are confirmed applicable to this case.}\\
\texttt{Include these hinted codes in the final <code> answer, then add any other applicable ICD-\{VERSION\} codes supported by the note.}\\[2pt]
\texttt{Hinted codes:}\\
\texttt{- \{CODE\_1\} (\{DESCRIPTION\_1\})}\\
\texttt{- \{CODE\_2\} (\{DESCRIPTION\_2\})}\\
\texttt{...}\\[2pt]
\texttt{Mini example:}\\
\texttt{If the hint says:}\\
\texttt{- 276.2 (Acidosis)}\\
\texttt{and the note also supports 401.9 and 584.9, then the final answer should include both the hinted code and the other supported codes:}\\
\texttt{<code>276.2, 401.9, 584.9</code>}
\end{quote}

Descriptions are taken from the ICD description files and truncated to a maximum length. The reward target remains the full gold set $Y_i$, not the hinted subset.

\section{Implementation Details}
\label{app:impl}

\paragraph{Hardware and runtime.}
Discriminative baselines are trained on a single A100 80GB GPU. Generative RL experiments are run on $4 \times$ H100 80GB GPUs. Stage-1 SFT takes about 24 hours. A Top-50 GRPO run takes about 120 hours, while a Full-code GRPO run takes about 160 hours. Each \method{} round adds one additional GRPO epoch initialized from the previous-round checkpoint.

\paragraph{Frameworks.}
Stage-1 SFT is implemented with LlamaFactory. Stages 2 and 3 use EasyR1 with vLLM-backed rollouts. The reward function, parser, and preprocessing utilities are released with the codebase to support reproducibility.

\paragraph{Long-note truncation.}
Discharge summaries can exceed the model context window. We therefore apply a tiered middle-out truncation scheme that prioritizes preserving section headers and the discharge-diagnoses block. This keeps high-value clinical structure and diagnosis anchors in the input while fitting long notes into the model context.

\section{Evaluation Metrics}
\label{app:metrics}

\paragraph{Definitions.}
For each predicted set $\hat{Y}$ and gold set $Y$ over the label vocabulary $\mathcal{C}$, we compute precision, recall, and F1 at both the micro and macro levels. Micro scores pool true positives, false positives, and false negatives across the full test set before computing each metric. Macro scores first compute per-class precision, recall, and F1, then report the arithmetic mean over classes following \citet{edin2023automated}. Macro-F1 is therefore the arithmetic mean of per-class F1, not the harmonic mean of macro precision and macro recall. This choice avoids the artificial deflation discussed in prior work.

\paragraph{Decision threshold tuning.}
For discriminative models, we tune a single decision threshold on the validation set to maximize micro-F1, then apply the same threshold at test time. We do not tune separate thresholds per class. Generative models do not use a decision threshold; their predictions are obtained by parsing the generated \texttt{<code>...</code>} span into a code set.

\section{LLM Usage}

We used Large Language Models (ChatGPT/Claude/Gemini) exclusively for grammatical correction in this manuscript. The LLMs played no role in research ideation, methodology, or scientific content generation. All technical contributions and scientific insights are original work by the authors.

\end{document}